\documentclass[sigconf, screen]{acmart}
\usepackage{tabularx}
\usepackage{multirow}
\usepackage{makecell}

\usepackage{graphicx}
\usepackage{subcaption} 
\usepackage[most]{tcolorbox} 
\newcommand{\xiaowu}{\fontsize{9pt}{9pt}\selectfont}        
\AtBeginEnvironment{tabularx}{\xiaowu}
\newcolumntype{Y}{>{\centering\arraybackslash}X}
\AtBeginDocument{%
  }

\definecolor{mygray}{rgb}{0.97,0.97,0.97}

\setcopyright{acmlicensed}
\copyrightyear{2025}
\acmYear{2025}
\acmDOI{XXXXXXX.XXXXXXX}
\acmConference[ACM XX 2025]{ In Proceedings of the ACM International Conference on XX}{June 03--05,
  2025}{Woodstock, NY}
\acmISBN{978-1-4503-XXXX-X/2025/04}

\begin{document}

\title{Talk is Not Always Cheap: Promoting Wireless Sensing Models with Text Prompts}

\author{Zhenkui Yang$^1$, Zeyi Huang$^2$, Ge Wang$^1$, Han Ding$^1$, Tony Xiao Han$^3$, Fei Wang$^{1}$}
\authornote{Ongoing porject, Fei Wang is the corresponding author.}
\affiliation{%
  \institution{$^1$ Xi'an Jiaotong University  \quad $^2$ University of Wisconsin - Madison \quad $^3$ Huawei Technologies Ltd.}
  \country{}}
\email{zkyang@stu.xjtu.edu.cn, 
zeyihuang@cs.wisc.edu, tony.hanxiao@huawei.com, {gewang,dinghan, feynmanw}@xjtu.edu.cn} 

\renewcommand{\shortauthors}{Yang et al.}

\begin{abstract}

     Wireless signal-based human sensing technologies, such as WiFi, millimeter-wave (mmWave) radar, and Radio Frequency Identification (RFID), enable the detection and interpretation of human presence, posture, and activities, thereby providing critical support for applications in public security, healthcare, and smart environments. These technologies exhibit notable advantages due to their non-contact operation and environmental adaptability; however, existing systems often fail to leverage the textual information inherent in datasets. To address this, we propose an innovative text-enhanced wireless sensing framework, WiTalk, that seamlessly integrates semantic knowledge through three hierarchical prompt strategies—label-only, brief description, and detailed action description—without requiring architectural modifications or incurring additional data costs. We rigorously validate this framework across three public benchmark datasets: XRF55 for human action recognition (HAR), and WiFiTAL and XRFV2 for WiFi temporal action localization (TAL). Experimental results demonstrate significant performance improvements: on XRF55, accuracy for WiFi, RFID, and mmWave increases by 3.9\%, 2.59\%, and 0.46\%, respectively; on WiFiTAL, the average performance of WiFiTAD improves by 4.98\%; and on XRFV2, the mean average precision gains across various methods range from 4.02\% to 13.68\%. Our codes have been included in  \url{https://github.com/yangzhenkui/WiTalk}.
\end{abstract}




\maketitle

\begin{figure}[H]
    \centering
    \includegraphics[width=\linewidth]{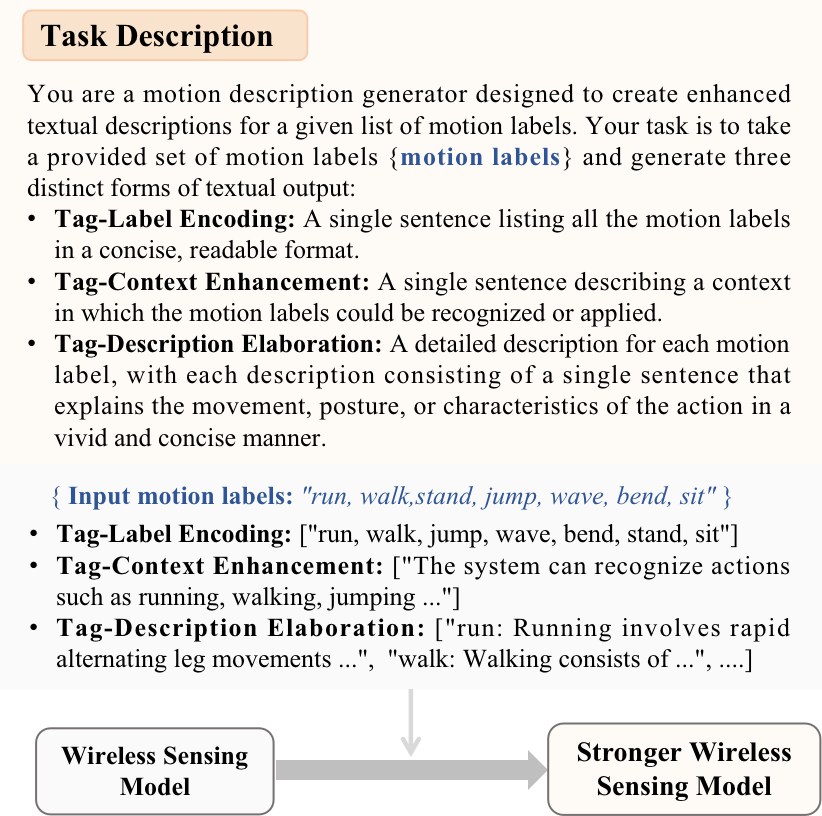}
    \caption{
    WiTalk transforms raw motion labels into hierarchical semantic representations. By leveraging large language models (LLMs), these enriched annotations enhance the performance of wireless sensing models for training and inference, enabling robust action recognition through integrated contextual semantics.
    }
    \label{fig:enter-label}
\end{figure}

\section{Introduction}\label{sec:introduction}

Human sensing, the capability to capture and interpret information about human presence, posture, or activities, serves as a foundational element for applications in public security, healthcare, smart homes, and human-computer interaction. Among various human sensing modalities, wireless signal-based approaches—particularly those utilizing Wi-Fi, millimeter-wave (mmWave) radar, and radio frequency identification (RFID) signals—have attracted considerable attention owing to their non-contact operation and strong resilience to environmental factors like lighting conditions and physical obstructions. In contrast, vision-based systems are constrained by privacy concerns and the need for a clear line of sight~\cite{mmfi}, while wearable sensors require user compliance and ongoing maintenance~\cite{imu2clip}, both of which fall short of the advantages offered by wireless sensing, which leverages ubiquitous wireless signals for passive detection. This characteristic makes wireless sensing particularly well-suited for real-world deployment.

Current research in wireless sensing often integrates multiple modalities, such as video, inertial measurement units (IMUs), and images, to enhance the expressive power of wireless signals. For instance, XRFV2~\cite{xrfmamba} combines Wi-Fi signals with IMU data, employing the XRFMamba network to capture long-term dependencies for action localization and summarization. Similarly, MM-Fi~\cite{mmfi} introduces the first multi-modal, non-intrusive 4D human dataset, comprising over 320,000 frames across five modalities, supporting tasks like action recognition and pose estimation. TS2ACT~\cite{ts2act} leverages semantically rich label text to generate an augmented image dataset, achieving few-shot human activity recognition (HAR) through cross-modal co-learning, with performance approaching or surpassing fully supervised methods. However, the independent collection of these multi-modal datasets increases costs and introduces challenges related to time synchronization and spatial calibration. Ensuring data consistency requires precise timestamp alignment and sensor positioning, often necessitating additional hardware. This leads to complex experimental setups, amplifying technical difficulties and workload.

Existing wireless sensing models predominantly focus on single-modal processing or vision-RF fusion, with text typically relegated to static labels rather than serving as a dynamic input to shape feature learning~\cite{fmfi, xrfmamba}. However, initial efforts have highlighted text’s potential. For example, TS2ACT and mmCLIP utilize text prompts to enable few-shot and zero-shot recognition, significantly improving data efficiency and model generalizability. Additionally, in more recent work, Wi-Chat~\cite{wi-chat} has preliminarily demonstrated the potential of integrating large language models (LLMs) with channel state information (CSI) signals, achieving zero-shot action recognition through prompt engineering, though this approach remains in its early stages. More notably, textual label data are inherently present in every dataset, with near-zero acquisition costs, offering a clear advantage over the complex collection processes required for Wi-Fi signals or IMU data. This underutilized potential has inspired our research.

Our work introduces an innovative approach, WiTalk, which significantly enhances human sensing capabilities based on wireless signals by seamlessly integrating a text branch into existing models. This method taps into the latent value of textual labels in datasets, improving the performance of wireless sensing models through feature fusion without altering their original architectures. Experimental results demonstrate that WiTalk achieves substantial performance gains across multiple datasets (XRF55~\cite{xrf55}, XRFV2~\cite{xrfmamba}, and WiFiTAL~\cite{wifitad}), fully validating its effectiveness. Specifically, on the XRF55 dataset, the incorporation of the text modality boosts accuracy by 3.9\%, 2.59\%, and 0.46\% for Wi-Fi, RFID, and millimeter-wave signals, respectively. On the WiFiTAL dataset, the WiFiTAD method with the text branch achieves an average performance improvement of 4.98\%. On the XRFV2 dataset, various methods incorporating the text branch show average mAP improvements ranging from 4.02\% to 13.68\%. These advancements underscore the critical role of the text branch in enhancing wireless action recognition.
In summary, our contributions can be highlighted in the following two aspects:



\noindent
\begin{itemize}
    \item We propose a method to enhance wireless sensing models using text labels from existing datasets, incurring no additional cost.
    \item We designed three hierarchical text prompt strategies with progressive semantic richness and systematically validated their performance in Wi-Fi, RFID, and mmWave action recognition and temporal action localization tasks across three large-scale public datasets.
\end{itemize}

\begin{figure*}[t]
    \centering
    \begin{subfigure}[b]{0.31\linewidth}
        \includegraphics[width=\linewidth]{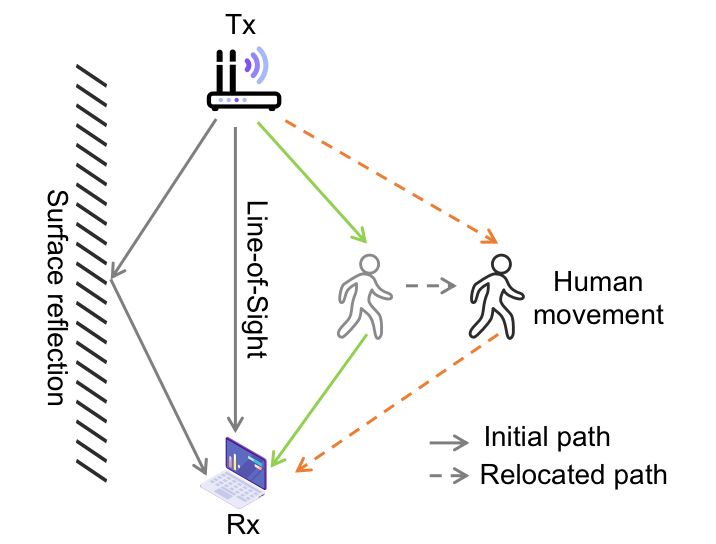}
        \caption{Human-Induced Multipath Effects}
        \label{fig:wifi}
    \end{subfigure}
    \hfill
    \begin{subfigure}[b]{0.31\linewidth}
        \includegraphics[width=\linewidth]{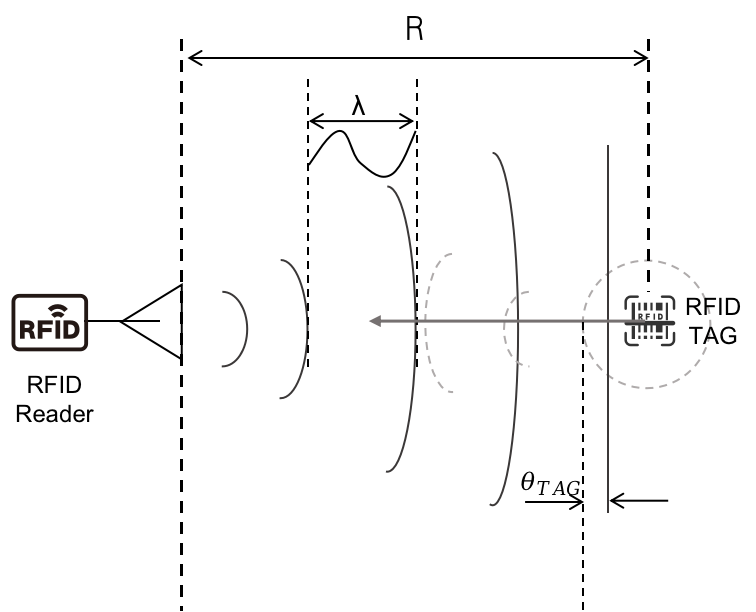}
        \caption{RFID System Operation Principle}
        \label{fig:rfid}
    \end{subfigure}
    \hfill
    \begin{subfigure}[b]{0.31\linewidth}
        \includegraphics[width=\linewidth]{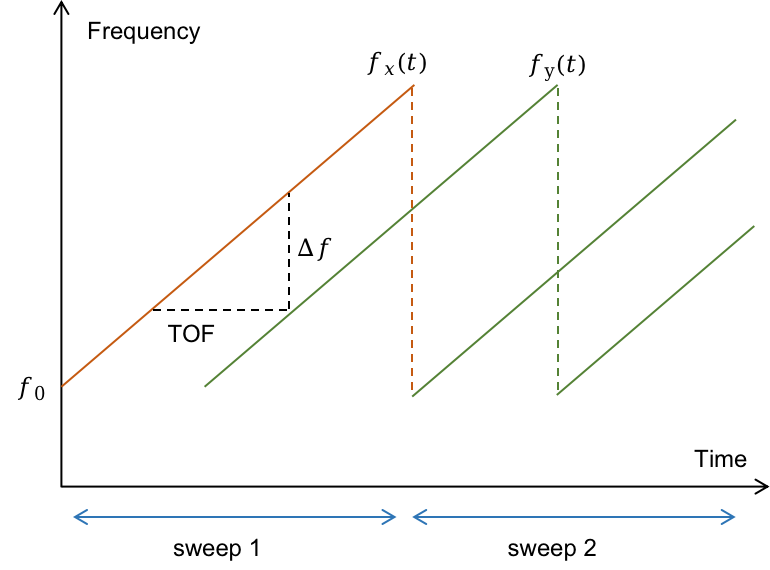}
        \caption{FMCW Operation}
        \label{fig:mmWave}
    \end{subfigure}
    \caption{The figure illustrates three fundamental signal propagation and sensing mechanisms used in wireless sensing systems; (a) Human-Induced Multipath Effects: Movement of the human body causes dynamic changes in multipath propagation, leading to variations in signal paths and reflections that can be captured by sensing systems; (b) RFID System Operation Principle: The RFID reader emits an wireless signal that is reflected back by the passive tag via backscatter communication. The signal's changes encode positional or identity information; (c) FMCW Operation: Frequency-Modulated Continuous Wave (FMCW) radar transmits chirp signals with linearly varying frequencies. The time delay between transmitted and received signals induces a frequency shift, which can be used to estimate target distance and motion through range-Doppler analysis.}
    \label{fig:pipeline-structure}
\end{figure*}

\section{Related Work}\label{sec:related-work}

\subsection{Wireless Sensing}\label{sec:wireless-sensing}

Wireless sensing leverages wireless signals such as Wi-Fi, mmWave, and RFID to achieve human activity recognition and related tasks. Compared to camera-based solutions, wireless sensing demonstrates significant advantages in indoor environments due to its ability to penetrate occlusions and preserve privacy.

\textbf{Wi-Fi Sensing:}  Wi-Fi signals, ubiquitous in indoor environments, are widely leveraged for human sensing research due to their reflective properties. A comprehensive survey \cite{wang2025survey} systematically categorizes Wi-Fi sensing techniques, encompassing applications such as indoor localization\cite{indoor}, identity recognition\cite{Turetta2022PracticalIR}, pose estimation \cite{Zhou2023MetaFiWT, yan2024person, Wang2019PersoninWiFiFP},  action recognition\cite{widar3.0} and health monitoring\cite{Yang2023RethinkingFD}. The mainstream approach employs Channel State Information (CSI) to capture fine-grained signal variations for activity recognition.  These studies highlight the immense potential of Wi-Fi sensing in non-intrusive perception, laying a robust foundation for further theoretical and practical advancements.

\textbf{mmWave Sensing:} mmWave signals enable high-resolution human motion recognition. EI~\cite{ei} combined Channel Impulse Response with Convolutional Neural Networks to achieve 65\% activity recognition accuracy, while SPARCS~\cite{sparcs} reached over 90\% ~ F1 scores through micro-Doppler spectrum analysis. RadHAR~\cite{radhar} and m-Activity~\cite{m-activity} achieved recognition accuracies of 90.47\% and 93.25\%, respectively, using point clouds, with real-time performance at 91.52 frames per second. PALMAR~\cite{palmar} achieved 91.88\% accuracy in multi-person activity tracking using clustering and domain adaptation techniques. In gesture recognition, Soli~\cite{soli} pioneered fine-grained gesture recognition with 92.1\% accuracy, while RFWash~\cite{rfwash} controlled gesture error rates below 8\% for hand hygiene monitoring. For handwriting tracking, mTrack~\cite{mtrack} achieved sub-millimeter (<8 mm) tracking errors, and mmWrite~\cite{mmwrite} and mmKey~\cite{mmkey} reached 95\% accuracy in character recognition tasks.
EgoHand~\cite{egohand} utilizes mmWave as a head-mounted wearable device for hand keypoint detection and hand gesture recognition, supporting VR and human-computer interaction scenarios.


\textbf{RFID Sensing:} RFID enables passive HAR via signal reflections for indoor positioning and gesture recognition. LANDMARC \cite{Jiang2009AnEA} uses reference tags for accurate, cost-effective indoor localization with active RFID. TagoramRT \cite{Yang2014TagoramRT} tracks mobile tags with high precision using phase values and a virtual antenna array. RFnet \cite{RFnet} employs CNN to recognize gestures and identities from RFID time-series signals, excelling in smart homes without heavy preprocessing. LoDiHAR \cite{LoDiHAR} offers a low-cost, distributed HAR system, using phase extraction and a GAN-Transformer model to achieve 94.9\% accuracy for eight activities at 10\% of COTS RFID costs.

In this paper, we leverage the Wi-Fi modality from the XRF55, WiFiTAL, and XRFV2 datasets to perform HAR and TAL tasks. 
Additionally, we utilize the mmWave and RFID modalities from the XRF55 dataset to further conduct HAR tasks.

\subsection{Language-Guided Action Understanding}

In recent years, an increasing number of studies have explored the integration of textual information using LLMs and prompt learning, providing new insights into wireless sensing tasks.

For example, Action-GPT~\cite{actiongpt} proposes a plug-and-play framework that uses LLMs to generate fine-grained action descriptions, enhancing the semantic alignment between text and motion representations. This framework supports a variety of model architectures and enables zero-shot motion generation. PLAR~\cite{plar} leverages multiple forms of prompts—including optical flow, vision models, and learnable prompts—to guide action recognition in videos, achieving notable accuracy improvements on datasets such as Okutamam (3.17–10.2\%). 
 ActionPrompt~\cite{actionprompt} introduces an Action Prompt Module that embeds textual labels and mines pose patterns to improve 2D-to-3D human pose estimation. STALE~\cite{stale} applies vision-language prompting to zero-shot temporal action detection, mitigating error propagation between proposal and classification stages. Meanwhile, mmCLIP~\cite{mmclip} extends this paradigm to mmWave signals by aligning them with textual semantics, enabling zero-shot human activity recognition in the wireless sensing domain.

Despite these advances, previous work has not explored the direct integration of textual semantics into wireless sensing systems—particularly without reliance on visual modalities or additional physical sensors.

\section{Preliminary}\label{sec:preliminary}

Wireless sensing leverages the intrinsic properties of wireless signals, such as reflection, penetration, and multipath propagation, to enable non-intrusive perception tasks like human action recognition and temporal action localization.
Unlike vision-based methods, wireless sensing operates effectively through occlusions and in varying lighting conditions, while preserving privacy by avoiding direct imaging.

\textbf{Principle of WiFi-Based Human Motion Sensing. }
Wi-Fi devices compliant with IEEE 802.11n/ac standards are generally equipped with multiple transmit and receive antennas, supporting Multiple-Input Multiple-Output (MIMO) technology. 
Each antenna pair between transmitter and receiver forms a MIMO channel comprising multiple subcarriers. 
These devices actively monitor the channel conditions to fine-tune transmission power and data rates, ensuring efficient use of the channel capacity. 
CSI describes channel characteristics, capturing each subcarrier’s frequency response of each subcarrier for every transmit-receive antenna pair. 
During Wi-Fi signal propagation, the received signal results from the constructive and destructive interference of multipath signals scattered by walls and surrounding objects. 
Human motions, such as hand movements, perturb the signal propagation paths, altering existing multipath effects or introducing new ones, as illustrated in Figure \ref{fig:wifi}, which depicts the multipath effects of Wi-Fi signals.
These changes are reflected in the CSI values and can be leveraged to detect and recognize human actions, such as gestures or posture changes.

Let \( M_T \) represent the number of transmit antennas, \( M_R \) represent the number of receive antennas, and \( S_c \) represent the number of OFDM subcarriers\cite{Ali2015KeystrokeRU}. Define \( X_i \) and \( Y_i \) as the \( M_T \)-dimensional transmitted signal vector and \( M_R \)-dimensional received signal vector, respectively, for subcarrier \( i \), and \( N_i \) as the \( M_R \)-dimensional noise vector. The MIMO system can be expressed by the following equation:
\begin{equation}
Y_i = H_i X_i + N_i \quad i \in [1, S_c]
\end{equation}
where the \( M_R \times M_T \)-dimensional matrix \( H_i \) represents the CSI for subcarrier \( i \). Wi-Fi devices estimate \( H_i \) by periodically transmitting known OFDM symbol preambles. 
As a human moves within the signal coverage area, variations in multipath effects cause fluctuations in CSI, capturing signal changes induced by motion and facilitating action detection.

\textbf{RFID Sensing Using Backscatter Communication. } 
RFID-based sensing leverages backscatter communication to enable interaction between a reader and passive tags, as illustrated in Figure \ref{fig:rfid}. 
The reader transmits Continuous Waves (CW) via the forward link, embedding commands such as Query or ACK to interrogate the tag. The passive tag, which does not possess an active radio transmitter, modulates its identification data (e.g., tag ID or response) onto the backscattered CW through the reverse link by varying its antenna impedance. 
This modulation process is powered by energy harvested from the incident CW\cite{Ding2017RFIPadEC}. The received power at the reader, denoted as $P_{rx}$, is determined by the radar equation, accounting for the two-way signal propagation:
\begin{equation}
P_{rx} = P_{tx} G_{tx} G_{rx} G_{tag}^2 \left(\frac{\lambda}{4\pi d}\right)^4
\label{eq:rfid}
\end{equation}
where $P_{tx}$ is the transmitted power, $G_{tx}$ and $G_{rx}$ are the transmit and receive antenna gains of the reader, respectively, $G_{tag}^2$ reflects the tag’s antenna gain for both forward and reverse signal paths., $\lambda$ is the wavelength of the CW, and $d$ is the distance between the reader and the tag. 
The term $\left(\frac{\lambda}{4\pi d}\right)^4$ reflects the two-way free-space path loss, as the signal travels from the reader to the tag and back. In addition to the tag’s identification data, the backscattered signal encapsulates environmental information through variations in $P_{rx}$. 
Perturbations in $P_{rx}$, induced by human presence or movement, alter the signal’s propagation path (e.g., via multipath effects or shadowing), enabling action recognition. By analyzing these variations, RFID sensing systems can detect and classify human activities in the deployment area.

\textbf{mmWave-Based Human Motion Sensing. } 
mmWave radar employs Frequency Modulated Continuous Wave (FMCW) technology to enable human motion sensing. 
As depicted in Figure~\ref{fig:mmWave}, the radar transmits a chirp signal with a linearly increasing frequency over time (illustrated by the orange dashed line \( f_x(t) \)). 
Upon reflection from a human body, the received signal (green solid line \( f_y(t) \)) exhibits a frequency shift \( \Delta f \) due to the Time-of-Flight (TOF). 
By mixing the transmitted and received signals, an Intermediate Frequency (IF) signal is obtained\cite{Adib20143DTV}. Applying a Fast Fourier Transform (FFT) to the IF signal allows the frequency offset \( \Delta f \) to be used for calculating the target distance:
\begin{equation}\label{eq:distance}
    R = \frac{c \cdot \Delta f \cdot T_c}{2B}
\end{equation}
where \( c \) is the speed of light, \( B \) is the frequency modulation bandwidth, and \( T_c \) is the chirp duration\cite{mmMesh}.

Across consecutive chirp cycles (e.g., sweep 1 and sweep 2 in the Figure~\ref{fig:mmWave}), human motion induces a Doppler effect, resulting in phase variations between adjacent chirps. A secondary FFT is performed to extract the Doppler shift and compute the velocity:
\begin{equation}\label{eq:velocity}
    v = \frac{\lambda \cdot \omega}{4\pi T_c}
\end{equation}
where \( \lambda \) is the wavelength of the millimeter wave, and \( \omega \) is the phase change rate. By integrating distance and Doppler data, the target angle can be further estimated using phase differences across an antenna array. The elevation angle \( \varphi \) is determined from the phase difference \( \omega_z \) between vertically aligned antennas, while the azimuth angle \( \theta \) is derived from the phase difference \( \omega_x \) between horizontally aligned antennas:
\begin{align}
    \varphi &= \sin^{-1}\left(\frac{\omega_z}{\pi}\right) 
    \label{eq:angle_phi}    \\
    \theta &= \sin^{-1}\left(\frac{\omega_x}{\cos\varphi \cdot \pi}\right)
    \label{eq:angle_theta}
\end{align}

Finally, the three-dimensional position of human joints is reconstructed by converting from spherical to Cartesian coordinates:
\begin{align}
    x &= R \cos\varphi \sin\theta
    \label{eq:coordinates_x}    \\
    z &= R \sin\varphi  
    \label{eq:coordinates_z}  \\
    y &= \sqrt{R^2 - x^2 - z^2} 
    \label{eq:coordinates_y}
\end{align}
This process enables high-precision capture and real-time analysis of subtle human movements.

These wireless signal characteristics form the technical foundation of our approach, supporting wireless sensing-based human activity recognition and temporal action localization.

\begin{figure}[t]
    \centering
    \includegraphics[width=1\linewidth]{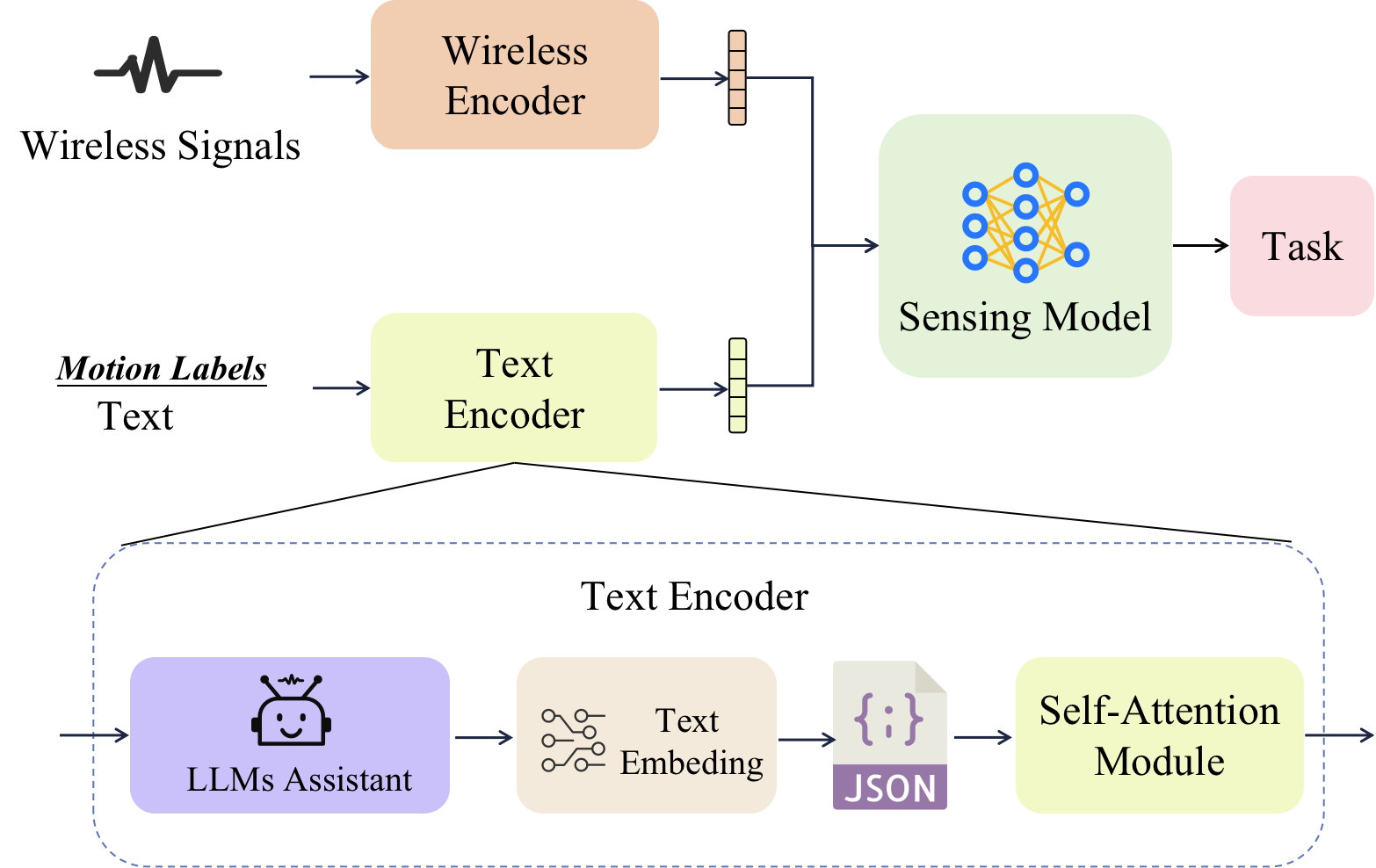}
    \caption{Overview of the WiTalk. Wireless signals are processed by a Wireless Encoder, while motion labels are enhanced by an LLMs Assistant and encoded into text embeddings. These embeddings are stored in JSON format, refined via an Attention Module, and fused with wireless features in the Sensing Model to perform downstream tasks.}
    \label{fig:method}
\end{figure}

\section{Methods}\label{sec:methods}

In this paper, we introduce a novel textual branch framework to improve wireless sensing-based action recognition (e.g., Wi-Fi, mmWave, RFID) by integrating multimodal fusion with textual semantics. Conventional wireless sensing models predominantly depend on spatiotemporal signal features, which restricts their ability to capture semantic context. To overcome this limitation, our framework incorporates a complementary textual modality, utilizing representations generated by LLMs to enrich the recognition process with deeper semantic understanding.

\subsection{Pipeline}\label{sec:pipeline}


The proposed WiTalk architecture is illustrated in Figure \ref{fig:method},  which depicts the information processing pipeline from input to output. Initially, we construct a contextual learning template (shown in Figure \ref{fig:enter-label}) to standardize textual label processing across datasets. These text labels are then semantically enhanced through LLMs. Subsequently, a text encoder transforms the enhanced textual labels into initial text feature vectors, which are further refined via an attention mechanism to improve their semantic expressiveness. These processed text feature vectors are then integrated with wireless signal feature vectors through weighted feature fusion. Finally, the fused feature vectors are fed into the sensing model to support downstream tasks and enable the training and optimization of multimodal action recognition.

Notably, text labels are encoded once and stored as key-value pairs in JSON files for efficient retrieval.

\subsection{Textual Prompts}\label{sec:prompt}

\subsubsection{Textual Feature Extraction and Label Processing. }
The textual branch begins with the extraction of textual features and processing of action labels, forming the cornerstone of our multimodal framework. To achieve this, we first define a contextual learning template to standardize the integration of textual labels across diverse datasets. This template provides a structured format that enriches raw action labels (e.g., "running," "walking") with consistent contextual cues, ensuring uniformity and adaptability in subsequent processing steps. The labels, paired with the contextual template, are then fed into an LLMs to perform semantic enhancement. This step leverages the LLM's natural language understanding capabilities to generate semantically enriched representations that capture deeper meanings and relationships beyond the original labels.

Next, the enhanced textual representations are processed by a text encoder, which transforms them into dense feature vectors. This encoding step converts the semantically augmented text into a numerical format suitable for downstream tasks, typically yielding vectors with a fixed dimensionality (e.g., 512 or 4096, depending on the encoder’s configuration). 

\subsubsection{Multi-Head Self-Attention Optimization}
To enhance the semantic expressiveness of \({T}_{\text{init}}\), particularly when multiple descriptions are involved (\(L > 1\)), we employ a Multi-Head Self-Attention (MHSA) mechanism. MHSA splits the input features into multiple subspaces and computes attention weights in parallel within each subspace, effectively capturing semantic relationships and contextual dependencies among descriptions. Specifically, MHSA first projects \({T}_{\text{init}}\) into Query, Key, and Value vectors via linear transformations. For each head \(h\), the scaled dot-product attention is computed as:
\begin{equation}
\text{Attention}(Q_h, K_h, V_h) = \text{Softmax}\left(\frac{Q_h K_h^\top}{\sqrt{d_k}}\right) V_h
\end{equation}
where \(Q_h = {T}_{\text{init}} W_h^Q\), \(K_h = {T}_{\text{init}} W_h^K\), and \(V_h = {T}_{\text{init}} W_h^V\) are the Query, Key, and Value vectors for the \(h\)-th head, with \(W_h^Q, W_h^K, W_h^V \in \mathbb{R}^{C \times d_k}\) as learnable projection matrices and \(d_k\) as the dimension per head. The outputs of all heads are then concatenated and transformed:
\begin{equation}
\text{MHSA}({T}_{\text{init}}) = \text{Concat}(\text{head}_1, \text{head}_2, \dots, \text{head}_H) W^O
\end{equation}
where \(W^O \in \mathbb{R}^{H d_k \times C}\) is the output projection matrix, yielding the enhanced representation:
\begin{equation}
{T}_{\text{att}} = \text{MHSA}({T}_{\text{init}}) + {T}_{\text{init}}    
\end{equation}

Inspired by mmCLIP, our approach leverages the MHSA mechanism to process the enhanced text features, further improving their contextual coherence and semantic richness. Subsequently, we concatenate \({T}_{\text{att}}\) with \({T}_{\text{init}}\) to form \({T}_{\text{combined}} \in \mathbb{R}^{B \times (L+1) \times C}\), thereby enhancing the robustness and contextual awareness of the textual features.

\subsubsection{Multimodal Fusion}
To imbue wireless signal features with semantic attributes, this section employs a weighted feature fusion approach to integrate textual and wireless signal features, with wireless signal features assigned a weight of 0.9 and text features a weight of 0.1. This weighting strategy underscores the primary role of wireless signal features in capturing spatiotemporal information for action recognition, while text features provide supplementary semantic support. The resulting fused feature vectors are then fed into a sensing model, providing robust support for various downstream tasks and enhancing performance in applications such as action recognition.

\subsection{Loss Function} \label{sec:loss function}
\subsubsection{Loss Function for TAL}
In Temporal Action Localization (TAL), the loss function optimizes both action classification and boundary localization. Features processed by a backbone network form a pyramid \( (f_1, f_2, \ldots, f_K) \), with each level \( f_i \) feeding into two branches for classification and localization predictions. The total loss is:

\begin{equation}
    L = \sum_{i=1}^K \left( \alpha_1 L_{\text{Cls},i} + \alpha_2 L_{\text{Loc},i} \right)
\end{equation}
where \( L_{\text{Cls},i} \) uses Focal Loss for classification, and \( L_{\text{Loc},i} \) employs a specialized loss for localization. Weights \( \alpha_1 = 1 \) and \( \alpha_2 = 1000 \) balance the two terms for effective training.

\subsubsection{Loss Function for HAR}

For Human Activity Recognition (HAR), Cross-Entropy Loss is used to assess multi-class classification performance. Given predicted probabilities \( \hat{y}_i \) and true labels \( y_i \), the loss is:

\begin{equation}
    L = \frac{1}{N} \sum_{i=1}^N \sum_{c=1}^C - y_{i,c} \log \hat{y}_{i,c}
\end{equation}
where \( N \) is the sample size, \( C \) is the number of classes, \( y_{i,c} \) is the one-hot encoded label, and \( \hat{y}_{i,c} \) is the predicted probability. This loss enhances classification accuracy for HAR’s temporal data.

\section{Evaluation}\label{sec:evaluation}

\subsection{Datasets}\label{sec:datasets}

\subsubsection{XRF55 Dataset. }
XRF55 comprises 42.9K samples from 39 subjects performing 55 indoor actions across five categories (Human-Object Interaction, Human-Human Interaction, Fitness, Body Motion, and Human-Computer Interaction). Data was collected in four scenes using Wi-Fi (Intel 5300 NIC, 200 Hz, 30 OFDM subcarriers), RFID (Impinj R420, 30 Hz), and mmWave radar (TI IWR6843ISK, 20 Hz), alongside synchronized Kinect videos. The dataset includes 30.0K training and 12.9K test samples, recorded in a rectangular sensing area with a 5-second action window, totaling 59h35min of multi-modal data. XRF55 was officially split with a 7:3 ratio into the training and testing sets.

\subsubsection{WiFiTAL Dataset. }
WiFiTAL consists of 553 untrimmed CSI recordings from three volunteers performing seven daily activities (walking, running, jumping, waving, falling, sitting, and standing). Data was collected in a $7\,m \times 12\,m \times 2.5\,m$ empty office environment using a single-antenna Intel 5300 NIC pair (TX/RX) operating on 30 OFDM sub-channels, sampled at 100\,Hz. The dataset contains 2,114 temporally segmented and annotated activity instances. WiFiTAL was officially split with a 7:3 ratio into the training and testing sets.

\subsubsection{XRFV2 Dataset. }
XRFV2 dataset contains synchronized multimodal recordings of 853 valid action sequences performed by 16 volunteers. Each volunteer conducted multiple variants of predefined action sequences across three indoor scenes (bedroom, study room, dining room), totaling 16 hours and 16 minutes of annotated data. The dataset includes Channel State Information (CSI) sampled from one transmitter (single antenna, 200 packets per second on channel 128 at 5.64\,GHz) and three receivers (each with three antennas, capturing CSI on 30 OFDM subcarriers), along with IMU data from wearable devices and RGB+D+IR videos. XRFV2 was officially split into training and testing sets with an 8:2 ratio, yielding 682 training and 171 testing sequences.

\subsection{Metrics}\label{sec:metrics}

\subsubsection{Evaluation Metrics for HAR}

Human Action Recognition (HAR) performance is typically measured using classification accuracy, which quantifies the proportion of correctly predicted actions among all action instances. Accuracy is computed as follows:

\begin{equation}
    \text{Accuracy} = \frac{\sum_{i=1}^{N} I(y_i = \hat{y}_i)}{N}
\end{equation}
where \( N \) denotes the total number of action instances, \( y_i \) and \( \hat{y}_i \) represent the ground-truth and predicted action labels respectively, and \( I \) is an indicator function outputting 1 if \( y_i = \hat{y}_i \), and 0 otherwise.

Higher accuracy values indicate better recognition capability and reliability of the model.

\subsubsection{Evaluation Metrics for TAL}

Temporal Action Localization (TAL) performance is primarily evaluated using temporal Intersection over Union (tIoU), extending Intersection over Union (IoU) from image detection to the temporal domain. 
Specifically, tIoU measures temporal overlap between predicted and ground-truth action segments:

\begin{equation}
    \begin{gathered}
    T_I = \max(0, \min(t_e, \hat{t}_e) - \max(t_s, \hat{t}_s)) \\
    T_U = \max(t_e, \hat{t}_e) - \min(t_s, \hat{t}_s) \\
    \text{tIoU} = \frac{T_I}{T_U}
    \end{gathered}
\end{equation}
where $t_s$, $t_e$ and $\hat{t}_s$, $\hat{t}_e$ denote the start and end times of ground-truth and predicted segments, respectively.

Predictions are deemed accurate when their tIoU scores surpass a designated cutoff (e.g., 0.5). 
To assess performance in greater detail, the Average Precision at threshold \( t \) (AP@t) is calculated using the following expression:

\begin{equation}
    AP@t = \frac{\sum_{i=1}^{N} I(\text{tIoU}_i \ge t)}{N}
\end{equation}
where \( N \) represents the total count of actions, and \( I \) serves as a binary indicator that yields 1 if \(\text{tIoU}_i \ge t\), and 0 otherwise.

Ultimately, the mean Average Precision (mAP) is derived by computing the mean of AP@t across various tIoU thresholds, offering a holistic evaluation of the TAL model's precision and resilience.

\subsection{Implementation Details}\label{sec:implementations}
\subsubsection{Experiments on the XRF55 Dataset. } For the experiments on the XRF55 dataset, the parameter settings adhered to the original paper. The batch size was set to 55, with a total of 200 training epochs. The initial learning rate was 0.0001, decaying by a factor of 0.5 every 40 epochs, and the Adam optimizer was employed. 
The training durations for each modality were as follows: approximately 8 hours for the WIF modality, 5 hours for the mmWave modality, and 3 hours for the RFID modality.

\subsubsection{Experiments on the WiFiTAL Dataset. } For the experiments on the WiFiTAL dataset, the Adam optimizer was used with an initial learning rate of 4e-5 and a weight decay coefficient of 1e-3. The batch size was set to 6, and training spanned 40 epochs.
with the total time from training to inference being approximately 6 hours. 
All other parameters remained consistent with the original paper.

\subsubsection{Experiments on the XRFV2 Dataset. } For the experiments on the XRFV2 dataset, the parameter settings largely follow those of the original paper. The batch size was set to 16, and all methods utilized the AdamW optimizer, with training conducted on a single Nvidia 3090 GPU for 80 epochs. The initial learning rate was set to 4e-5, decaying by a factor of 0.5 every 30 epochs. The Mamba version used was 1.0.1, and the causal\_conv1d version was 1.0.0.
The time required for each method, from training to inference, ranged approximately between 10 and 25 hours, varying depending on the specific method but consistently falling within this range.

To ensure fairness, all experiments across each dataset in this paper were conducted under unified parameter settings.

\begin{figure*}[t]
    \centering
    \includegraphics[width=1\linewidth]{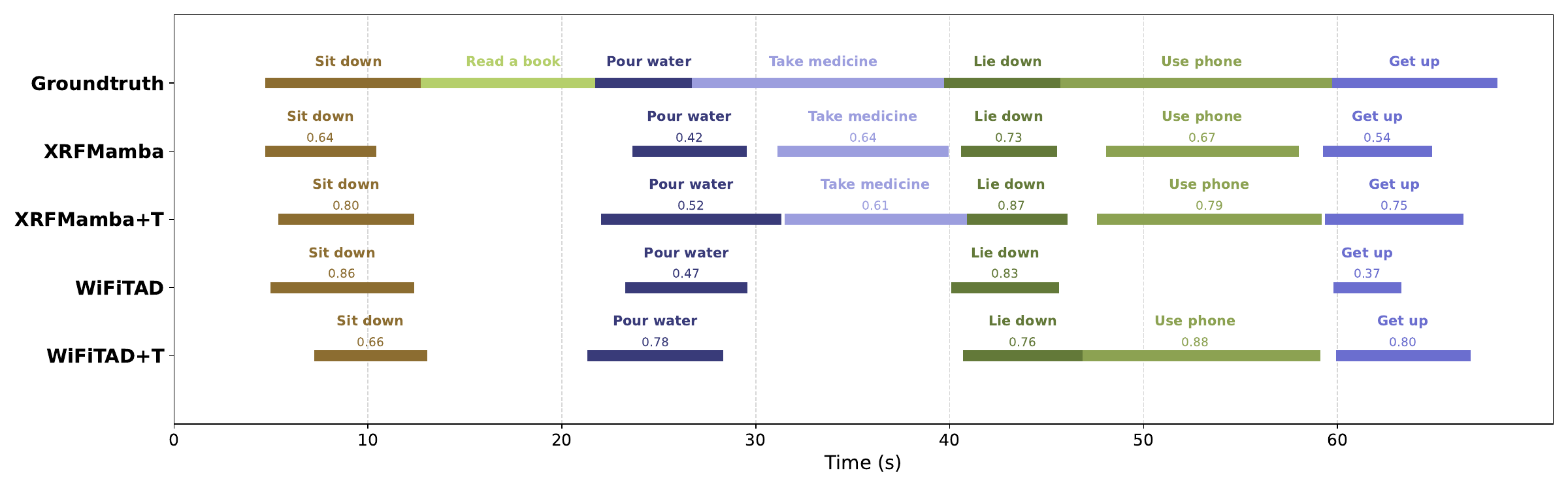}
    \caption{Visualization examples of the performance of XRFMamba and WiFiTAD on the XRFV2 dataset, where results marked with +T denote those obtained after incorporating the text modality.}
    \label{fig:visual_example}
\end{figure*}

\subsection{Results}\label{sec:results}
All results in this section are derived from text features encoded using the CLIP encoder, a choice that ensures high reliability and consistency during multimodal fusion. 


Experimental results show that our proposed method achieves significant performance improvements across multiple datasets , validating its versatility and practicality. Here, $W$ denotes the performance of the original wireless-only method, while $W+T$ represents the performance after integrating the text branch.

On the XRF55 dataset, Table~\ref{tab: xrf55_res} indicates that incorporating the text modality improves the accuracy of Wi-Fi, RFID, and mmWave by 3.9\%, 2.59\%, and 0.46\%, respectively.
These improvements, observed across heterogeneous wireless modalities, indicate that semantic textual prompts can complement different signal types effectively, regardless of their frequency band or sensing characteristics.


On the WiFiTAL dataset, Table~\ref{tab: wifital_res} reveals that the WiFiTAD model, after integrating the text branch, achieves mAP improvements ranging from 3.81\% to 5.88\% across various tIoU thresholds, with an average gain of 4.98\%, underscoring its stable and reliable enhancement in temporal tasks.

On the XRFV2 dataset, Table~\ref{tab: xrfv2_res} demonstrates that evaluations of various methods using the original paper’s code (e.g., XRFMamba, WiFiTAD, and UWiFiAction) show the text modality boosts mAP by an average of 4.02\% to 13.68\%, with particularly notable gains in TriDet and XRFMamba at 13.68\% and 13.19\%, respectively.


These experimental results strongly substantiate the critical role of the text branch in enhancing wireless signal-based action recognition capabilities. 
Furthermore, Figure~\ref{fig:visual_example} provides intuitive visualization examples, showcasing the performance improvements of the XRFMamba and WiFiTAD methods on the XRFV2 dataset after integrating the text branch.


\begin{table}[t]
    \renewcommand{\arraystretch}{1.2}
    \centering
    \caption{Accuracy Comparison of DML Across Modalities on XRF55}
    \label{tab: xrf55_res}
    \begin{tabularx}{0.47\textwidth}{c l *{3}{Y}}
        \toprule
        \multirow{2}{*}{\textbf{Model}} & \multirow{2}{*}{\textbf{Modal}} & \multicolumn{3}{c}{\textbf{Accuracy}} \\
        \cmidrule(lr){3-5}
        & & Wi-Fi & RFID & mmWave\\
        \midrule
        \multirow{3}*{DML \cite{xrf55}} & W & 87.26 & 56.82 & 86.88 \\
        & W+T & 91.16 & 59.41 & 87.34 \\
        & $\Delta$ $\uparrow$ & \textcolor{teal}{3.9} &  \textcolor{teal}{2.59} & \textcolor{teal}{0.46}  \\
        \bottomrule
    \end{tabularx}
\end{table}

\begin{table}[t]
    \renewcommand{\arraystretch}{1.2}
    \centering
    \caption{Performance Comparison of WiFiTAD Across Modalities on WiFiTAL}
    \label{tab: wifital_res}
    \begin{tabularx}{0.47\textwidth}{c l *{5}{Y} X} 
        \toprule
        \multirow{2}{*}{\textbf{Model}} & \multirow{2}{*}{\textbf{Modal}} & \multicolumn{5}{c}{\textbf{mAP @ tIoU}} & \multirow{2}{*}{\textbf{Avg}} \\
        \cmidrule(lr){3-7} 
        & & 0.3 & 0.4 & 0.5 & 0.6 & 0.7 & \\
        \midrule
        \multirow{3}{*}{WiFiTAD \cite{wifitad}} & W & 79.31 & 74.91 & 69.82 & 61.56 & 45.96 & 66.31 \\
         & W+T & 83.12 & 80.79 & 75.62 & 66.18 & 50.72 & 71.29 \\
        & $\Delta$ $\uparrow$ & \textcolor{teal}{3.81} & \textcolor{teal}{5.88} & \textcolor{teal}{5.80} & \textcolor{teal}{4.62} & \textcolor{teal}{4.76} & \textcolor{teal}{4.98} \\
        \bottomrule
    \end{tabularx}
    \label{tab:new_model_comparison}
\end{table}

\begin{table}[t]
    \renewcommand{\arraystretch}{1.2}
    \centering 
    \caption{Performance Comparison of Methods on XRFV2}
    \label{tab: xrfv2_res}
    \begin{tabularx}{0.5\textwidth}{c l *{6}{Y}} 
        \toprule
        \multirow{2}*{\textbf{Model}} & \multirow{2}*{\textbf{Modal}} & \multicolumn{5}{c}{\textbf{mAP @ tIoU}} & \multirow{2}*{\textbf{Avg}} \\
        \cmidrule(lr){3-7}
        & & 0.5 & 0.55 & 0.6 & 0.65 & 0.7 & \\
        
        \midrule  
        \multirow{3}*{TemporalMaxer \cite{temporalMaxer}} & W & 44.62 & 40.38 & 33.75 & 28.43 & 21.74 & 33.78 \\
        & W + T & 55.48 & 48.71 & 41.23 & 32.55 & 22.52 & 40.10 \\
        & $\Delta$ $\uparrow$ & \textcolor{teal}{10.86} & \textcolor{teal}{8.33} & \textcolor{teal}{7.48} & \textcolor{teal}{4.12} & \textcolor{teal}{0.78} & \textcolor{teal}{6.32} \\
        
        \midrule
        \multirow{3}*{ActionFormer \cite{actionformer}} & W & 55.59 & 51.48 & 45.51 & 37.72 & 25.40 & 43.14 \\
        & W + T & 65.07 & 59.96 & 51.48 & 42.06 & 31.18 & 49.95 \\
        & $\Delta$ $\uparrow$ & \textcolor{teal}{9.48} & \textcolor{teal}{8.48} & \textcolor{teal}{5.97} & \textcolor{teal}{4.34} & \textcolor{teal}{5.78} & \textcolor{teal}{6.81} \\
        
        \midrule   
        \multirow{3}*{TriDet \cite{tridet}} & W & 47.68 & 44.18 & 37.94 & 27.00 & 19.30 & 35.22 \\
        & W + T & 62.94 & 58.00 & 51.37 & 41.81 & 30.39 & 48.90 \\
        & $\Delta$ $\uparrow$ & \textcolor{teal}{15.26} & \textcolor{teal}{13.82} & \textcolor{teal}{13.43} & \textcolor{teal}{14.81} & \textcolor{teal}{11.09} & \textcolor{teal}{13.68} \\
        
        \midrule 
        \multirow{3}*{WiFiTAD \cite{wifitad}} & W & 62.90 & 58.97 & 53.47 & 45.87 & 36.56 & 51.55 \\
        & W + T & 67.41 & 64.12 & 59.00 & 49.90 & 40.15 & 56.12 \\
        & $\Delta$ $\uparrow$ & \textcolor{teal}{4.51} & \textcolor{teal}{5.15} & \textcolor{teal}{5.53} & \textcolor{teal}{4.03} & \textcolor{teal}{3.59} & \textcolor{teal}{4.57} \\

        \midrule  
        \multirow{3}*{UWiFiAction \cite{ushape}} & W & 43.05 & 39.89 & 34.20 & 27.59 & 22.13 & 33.37 \\
        & W + T & 48.02 & 43.54 & 37.69 & 32.44 & 25.27 & 37.39 \\
        & $\Delta$ $\uparrow$ & \textcolor{teal}{4.97} & \textcolor{teal}{3.65} & \textcolor{teal}{3.49} & \textcolor{teal}{4.85} & \textcolor{teal}{3.14} & \textcolor{teal}{4.02} \\
        
        \midrule   
        \multirow{3}*{ActionMamba \cite{actionmamba}} & W & 70.97 & 68.55 & 63.96 & 58.26 & 47.65 & 61.88 \\
        & W + T & 75.54 & 73.56 & 69.28 & 63.00 & 55.09 & 67.29 \\
        & $\Delta$ $\uparrow$ & \textcolor{teal}{4.57} & \textcolor{teal}{5.01} & \textcolor{teal}{5.32} & \textcolor{teal}{4.74} & \textcolor{teal}{7.44} & \textcolor{teal}{5.42} \\
        
        \midrule   
        \multirow{3}*{XRFMamba \cite{xrfmamba}} & W & 65.40 & 60.51 & 54.67 & 47.75 & 34.70 & 52.61 \\
        & W + T & 72.95 & 70.74 & 67.61 & 63.43 & 54.25 & 65.80 \\
        & $\Delta$ $\uparrow$ & \textcolor{teal}{7.55} & \textcolor{teal}{10.23} & \textcolor{teal}{12.94} & \textcolor{teal}{15.68} & \textcolor{teal}{19.55} & \textcolor{teal}{13.19} \\
        \bottomrule
    \end{tabularx}
\end{table}

\subsection{Ablation Study}\label{sec:ablation-study}

The key to text-augmented wireless sensing lies in aligning signal features with semantic descriptions. To this end, we conduct a systematic evaluation of the effects of prompt engineering and embedding model selection, comparing three advanced text encoders: CLIP-ViT-Large-Patch14 (CLIP)\cite{clip}, LLaMA-2-7B-Chat-HF (LLaMA)\cite{llama2}, and Text-Embedding-V3 (Qwen)\cite{qwen}, with three prompt strategies: TLE (Tag-Label Encoding), TCE (Tag-Context Enhancement), and TDE (Tag-Description Elaboration). 
Notably, the baselines reported in the tables refer to the original method without text augmentation.

On the XRF55 dataset, Table~\ref{tab:as_xrf55_res} demonstrates that CLIP and LLaMA achieve superior performance for Wi-Fi and mmWave modalities. However, LLaMA’s RFID accuracy falls below the baseline, likely because its dialogue-oriented embeddings struggle to capture RFID’s action-specific patterns. On the WiFiTAL dataset, Table \ref{tab:as_wifital_res} indicates CLIP with TCE achieves the highest average mAP of 71.29\%, with TDE excelling at higher tIoU thresholds. On the XRFV2 dataset, Table \ref{tab:as_xrfv2_res} reveals CLIP with TCE reaches an average mAP of 65.80\%, outperforming Qwen and LLaMA. Overall, CLIP proves most robust, with TCE and TDE enhancing semantic expressiveness.

On XRFV2, Qwen’s TCE and TDE score only 9.36\% and 11.62\% respectively, and LLaMA’s TCE is 13.74\%. This weaker performance may arise from a mismatch with the dataset’s semantic needs. Qwen, designed for general text, struggles with action-specific semantics, while LLaMA’s dialogue focus limits its alignment with wireless signal spatiotemporal features.

\begin{table}[t]
    \renewcommand{\arraystretch}{1.2}
    \centering
    \caption{Action Classification Accuracy Comparison of Embedding Models and Prompt Strategies on XRF55 Dataset}
    \label{tab:as_xrf55_res}
    \begin{tabularx}{0.47\textwidth}{c c *{3}{Y}}
        \toprule
        \multirow{2}{*}{\textbf{\makecell{Embedding\\Model}}} & \multirow{2}{*}{\textbf{\makecell{Text\\Strategy}}} & \multicolumn{3}{c}{\textbf{Accuracy}} \\
        \cmidrule(lr){3-5}
        & & Wi-Fi & RFID & mmWave\\
        \midrule
        baseline & - & 87.26 & 56.82 & 86.88 \\
        \midrule
        \multirow{3}{*}{\textbf{CLIP}} 
            & TLE & 87.99 & 55.16 & 86.91  \\
            & TCE & \textbf{91.16} & 59.41 & 87.34  \\
            & TDE & 88.61 & \textbf{60.13} & 87.77 \\
        \midrule
         \multirow{3}{*}{\textbf{Qwen}} 
            & TLE & 87.70 & 58.43 & 86.28  \\
            & TCE & 87.61 & 58.96 & 87.19  \\
            & TDE & 87.93 & 53.26 & 87.67 \\
        \midrule
         \multirow{3}{*}{\textbf{LLaMA}} 
            & TLE & 88.21 & 52.94 & 86.45  \\
            & TCE & 89.22 & 50.98 & \textbf{88.78}  \\
            & TDE & 87.66 & 53.62 & 87.38 \\
        \bottomrule
    \end{tabularx}
\end{table}

\begin{table}[t]
    \renewcommand{\arraystretch}{1.2}
    \centering
    \caption{Effect of Textual Embedding Strategies on WiFiTAD’s mAP Across tIoU Thresholds on WiFiTAL}
    \label{tab:as_wifital_res}
    \begin{tabularx}{0.47\textwidth}{c c *{5}{Y} c} 
        \toprule
        \multirow{2}{*}{\textbf{\makecell{Embedding\\Model}}} & \multirow{2}{*}{\textbf{\makecell{Text\\Strategy}}} & \multicolumn{5}{c}{\textbf{mAP @ tIoU}} & \multirow{2}{*}{\textbf{Avg}} \\
        \cmidrule(lr){3-7}
        & & 0.3 & 0.4 & 0.5 & 0.6 & 0.7 & \\
        \midrule
        baseline & - & 79.31 & 74.91 & 69.82 & 61.56 & 45.96 & 66.31 \\
        \midrule
        \multirow{3}{*}{\textbf{CLIP}} 
            & TLE & 80.32 & 78.29 & 74.18 & 66.78 & 48.06 & 69.73 \\
            & TCE & 83.12 & 80.79 & 75.62 & 66.18 & \textbf{50.72} & \textbf{71.29} \\
            & TDE & \textbf{84.23} & \textbf{81.92} & 75.04 & 66.24 & 44.97 & 70.48 \\
        \midrule
        \multirow{3}{*}{\textbf{Qwen}} 
            & TLE & 77.11 & 74.80 & 71.26 & 61.28 & 44.10 & 65.71 \\
            & TCE & 77.86 & 75.35 & 71.43 & 64.62 & 44.29 & 66.71 \\
            & TDE & 79.74 & 76.03 & 71.29 & 64.76 & 47.91 & 67.95 \\
        \midrule
        \multirow{3}{*}{\textbf{LLaMA}} 
            & TLE & 77.54 & 73.92 & 70.45 & 64.67 & 47.93 & 66.90 \\
            & TCE & 83.17 & 79.58 & 72.88 & 64.55 & 45.97 & 69.23 \\
            & TDE & 81.96 & 77.63 & \textbf{75.71} & \textbf{69.42} & 50.53 & 71.05 \\        
        \bottomrule
    \end{tabularx}
    \label{tab:text_embedding_impact}
\end{table}

\begin{table}[t]
    \renewcommand{\arraystretch}{1.2}
    \centering 
    \small
    \caption{Effect of Textual Embedding Strategies on XRFMamba’s mAP Across tIoU Thresholds on XRFV2}
    \label{tab:as_xrfv2_res}
    \begin{tabularx}{0.47\textwidth}{c c *{6}{Y}} 
        \toprule
        \multirow{2}*{\textbf{\makecell{Embedding\\Model}}} & \multirow{2}*{\textbf{\makecell{Text\\Strategy}}} & \multicolumn{5}{c}{\textbf{mAP @ tIoU}} & \multirow{2}*{\textbf{Avg}} \\
        \cmidrule(lr){3-7}
        & & 0.5 & 0.55 & 0.6 & 0.65 & 0.7 & \\
        \midrule   
        baseline & - & 65.40 & 60.51 & 54.67 & 47.75 & 34.70 & 52.61 \\
        \midrule
        \multirow{3}*{\textbf{CLIP}} 
            & TLE & 66.33 & 61.22 & 54.98 & 47.22 & 37.93 & 53.54 \\
            & TCE & \textbf{72.95} & \textbf{70.74} & \textbf{67.61} & \textbf{63.43} & \textbf{54.25} & \textbf{65.80} \\
            & TDE & 69.33 & 66.87 & 59.77 & 51.85 & 43.24 & 58.21 \\
        \midrule
        \multirow{3}*{\textbf{Qwen}} 
            & TLE & 65.24 & 62.34 & 56.97 & 49.15 & 39.49 & 54.64 \\
            & TCE & 16.65 & 13.21 & 9.97 & 4.54 & 2.42 & 9.36 \\
            & TDE & 21.34 & 15.79 & 10.28 & 6.64 & 4.05 & 11.62 \\
        \midrule
        \multirow{3}*{\textbf{LLaMA}} 
            & TLE & 64.08 & 59.00 & 52.86 & 47.59 & 39.15 & 52.54 \\
            & TCE & 21.18 & 17.37 & 13.48 & 9.71 & 6.95 & 13.74 \\
            & TDE & 63.32 & 58.64 & 52.09 & 44.17 & 36.10 & 50.86 \\
        \bottomrule
    \end{tabularx}
\end{table}

\section{Discussion and Limitation}\label{sec:discussion}


This paper validates the potential of integrating a text branch into a wireless signal-based human sensing framework for applications in HAR and TAL tasks. However, this approach still exhibits several notable limitations. Firstly, the performance improvements brought by the text branch vary inconsistently across different types of wireless signals and datasets. For instance, on the XRF55 dataset, the accuracy of Wi-Fi signals improved by 3.9\%, whereas mmWave signals saw only a marginal increase of 0.46\%. This suggests that the enhancement effect of text features may be constrained by the characteristics of the signal modality, depending on data quality and signal type in specific scenarios. Secondly, the method heavily relies on the quality of text descriptions and the expressive capacity of the embedding model. If the semantic richness of the text prompt is insufficient the full guiding potential of this approach may not be realized. 
The design of an optimal prompt strategy requires further exploration 
and refinement.

Moreover, the underlying mechanism by which encoded text features enhance the precision of wireless sensing remains unclear. For example, the CLIP model, trained solely on text and image data without exposure to wireless signals, can still significantly boost wireless sensing performance—an intriguing phenomenon that warrants deeper investigation.

\section{Conclusion}\label{sec:conclusion}
In this work, we introduce a novel framework that enhances wireless signal-based human sensing by integrating a textual branch, leveraging text semantics to improve signal processing. 
When integrated with existing models, our framework achieves performance gains without requiring structural changes, highlighting the framework’s flexibility. Evaluations show that combining advanced embedding models with detailed text descriptions yields optimal results, laying a foundation for text-augmented wireless sensing.

Our findings reveal text semantics as a low-cost, effective complement to wireless signals, linking signal processing to contextual understanding. 
This advance extends wireless signal-based perception and supports privacy-preserving multimedia systems.

\bibliographystyle{ACM-Reference-Format}
\bibliography{sample-base}

\end{document}